# Comparisons between a Large Language Model-based Real-Time Compound Diagnostic Medical AI Interface and Physicians for Common Internal Medicine Cases using Simulated Patients

Hyungjun Park, M.D., Ph.D.[1,2*], Chang-Yun Woo, M.D.[3*], Seungjo Lim[2], Seunghwan Lim [2], Keunho Kwak[2], Ju Young Jeong[4], Chong Hyun Suh, M.D., Ph.D.[4]

[1]Department of Pulmonology, Shihwa Medical Center, Siheung, Republic of Korea

[2]Helpmedoc Inc., Republic of Korea

[3]Department of Internal Medicine, Asan Medical Center, Seoul, Republic of Korea

[4]Department of Radiology and Research Institute of Radiology, University of Ulsan College of Medicine, Asan Medical Center, Seoul, Republic of Korea

[*]Both contributed equally.

**Address correspondence to:**

Chong Hyun Suh, M.D., Ph.D.

Department of Radiology and Research Institute of Radiology, Asan Medical Center,

University of Ulsan College of Medicine, Seoul, Republic of Korea

Tel: 82-2-3010-1648

Fax: 82-2-476-0090

E-mail: chonghyunsuh@amc.seoul.kr

**Funding:** No

**Conflicts of interest:** No

**Type of manuscript:** Original investigation

**Word count for text:** 2527

**Key Points**

**Question** Does an LLM-based, real-time, compound diagnostic medical AI interface help with common internal medicine cases based on United States Medical Licensing Examination Step 2 Clinical Skills style exams?

**Findings** In a non-randomized clinical trial including 10 clinical vignettes, the accuracy of the first differential diagnosis was within the range of 50–70% for three physicians and 80% for the AI interface, while the interface required 44.6% less time, reduced costs by 98.1%, and achieved comparable patient satisfaction scores.

**Meaning** In a clinical trial involving first-time patients in primary care consultations for common internal medicine cases, the AI interface achieved results in less time and at a lower cost than physicians, demonstrating a relatively higher diagnostic accuracy and comparable patient satisfaction.

**Abstract**

**Objective** To develop an LLM based realtime compound diagnostic medical AI interface and performed a clinical trial comparing this interface and physicians for common internal medicine cases based on the United States Medical License Exam (USMLE) Step 2 Clinical Skill (CS) style exams.

**Methods** A nonrandomized clinical trial was conducted on August 20, 2024. We recruited one general physician, two internal medicine residents (2nd and 3rd year), and five simulated patients. The clinical vignettes were adapted from the USMLE Step 2 CS style exams. We developed 10 representative internal medicine cases based on actual patients and included information available on initial diagnostic evaluation. Primary outcome was the accuracy of the first differential diagnosis. Repeatability was evaluated based on the proportion of agreement.

**Results** The accuracy of the physicians' first differential diagnosis ranged from 50% to 70%, whereas the realtime compound diagnostic medical AI interface achieved an accuracy of 80%. The proportion of agreement for the first differential diagnosis was 0.7. The accuracy of the first and second differential diagnoses ranged from 70% to 90% for physicians, whereas the AI interface achieved an accuracy rate of 100%. The average time for the AI interface (557 sec) was 44.6% shorter than that of the physicians (1006 sec). The AI interface ($0.08) also reduced costs by 98.1% compared to the physicians' average ($4.2). Patient satisfaction scores ranged from 4.2 to 4.3 for care by physicians and were 3.9 for the AI interface

**Conclusion** An LLM based realtime compound diagnostic medical AI interface demonstrated diagnostic accuracy and patient satisfaction comparable to those of a physician, while requiring less time and lower costs. These findings suggest that AI interfaces may have the potential to assist primary care consultations for common internal medicine cases.

**Introduction**

Healthcare disparities are becoming increasingly pronounced. Access to medical services varies significantly depending on the country and even between urban and rural areas, making it challenging for patients to receive care easily.[1] Scheduling appointments, waiting for consultations, and incurring costs add to the burden.[2] Patients who are uncertain about whether to seek medical attention often turn to the Internet for information.[3,4] However, accurate high-quality medical information is often difficult to obtain.

Large language models (LLMs) are advanced artificial intelligence (AI) systems designed to process input requests and generate responses in natural language. A recent study of healthcare workers highlighted initial concerns about credibility and reliability in AI chatbots like ChatGPT.[5] Despite these concerns, the study also emphasized the potential benefits of AI in healthcare, such as enhancing efficiency, supporting decision-making, and improving accessibility. Furthermore, research has shown that LLMs can reduce physician workload while maintaining or even improving consistency in clinical responses.[6] Moreover, advancements in reasoning capabilities and diagnostic accuracy have increasingly demonstrated the potential for such systems to augment clinical workflows effectively.[7] LLMs are being increasingly integrated into various healthcare settings owing to their versatility and broad applicability.[8,9] Several recent studies have explored the impact of LLM-based chatbots on medical decisions.[10,11] However, while these studies have introduced models capable of conducting diagnostic consultations, they have not yet been implemented as real-world services.[12] To date, no studies have examined an LLM-based real-time compound diagnostic medical AI interface. Moreover, insufficient transparency in stochasticity reporting in LLM studies for medical applications has been reported.[13,14]

We developed an LLM-based real-time compound diagnostic medical AI interface and performed a clinical trial comparing this interface and physicians for common internal medicine cases based on the United States Medical License Exam Step 2 Clinical Skill style exams.

**Methods**

This study was reviewed and determined to be exempt from institutional review board approval at Asan Medical Center. This study was conducted based on the Minimum Reporting Items for Clear Evaluation of Accuracy Reports of Large Language Models in Healthcare (MI-CLEAR-LLM) checklist.[15,16]

*Study Design*

A non-randomized clinical trial was conducted on August 20, 2024. We recruited one general physician and two internal medicine residents, with one in their second year and the other in their third year, through email lists at Asan Medical Center. In addition, we recruited five simulated patients who had experience as standardized patients.

Five simulated patients were each randomly assigned two clinical vignettes with which they familiarized themselves beforehand. Each simulated patient engaged in Q&A-style chats regarding the clinical vignettes with three physicians and a real-time compound diagnostic medical AI interface (Figure 1A). The physicians, who were blinded to the clinical vignettes, interacted with all five simulated patients, experiencing 10 clinical vignettes in total (Figure 1B). A moderator created three simultaneous chat rooms for the physicians and facilitated the entry and exit of the simulated patients. One-to-one real-time chats between physicians and simulated patients were conducted using a computer-based, commercialized instant messaging system, and the recommended session length was 15 minutes per patient. The physicians provided three differential diagnoses after conducting real-time chats with the simulated patients.

Simultaneously, the simulated patients accessed https://www.hmdoc.kr and engaged in real-time chats with a real-time compound diagnostic medical AI interface using the same clinical vignettes. After the Q&A session, the real-time compound diagnostic medical AI interface also provided three differential diagnoses. To assess repeatability, the process was performed again in three distinct sessions

*Clinical Vignettes*

The clinical vignettes were adapted from the United States Medical License Exam Step 2 Clinical Skill style exams. We developed 10 representative internal medicine cases based on actual patients and included information available on initial diagnostic evaluation, including chief complaint, vital signs, detailed history of present illness, medical history, and physical examination. A representative clinical vignette is included in Supplementary materials. For each clinical vignette, two internal medicine physicians (C.Y.W., 10 years of experience in endocrinology, and H.P., 8 years of experience in pulmonology) determined the three most likely differential diagnoses as a reference standard by consensus.

*Instructions for Physicians and Simulated Patients*

The instructions for physicians were as follows: "The physician conducts an initial consultation with a new patient through chat for approximately 10 minutes. At the beginning of the conversation, the patient provides the doctor with their vital signs, which should be referenced during the consultation. The physician provides up to three differential diagnoses in the order of likelihood based on the consultation results. For the highest-priority diagnosis, the doctor explains a brief lifestyle guideline and treatment plan. The process is similar to that of a typical medical consultation. After receiving an explanation of the differential diagnosis and treatment plan, the patient can ask additional questions. Each conversation is limited to a maximum of 15 minutes per clinical vignette."

*Development of LLM-based Real-Time Compound Diagnostic Medical AI Interface*

This study introduced and evaluated a medical AI agent that operates in a chat-based format to assist in patient consultations (Figure 2). This system first deploys a Basic History LLM, which asks 10–12 initial questions to collect core information about the patient's concerns. Based on the patients' responses, the agent decides whether to link the patient to a predefined questionnaire (Detailed Symptom Survey) or proceed with further interactive questioning (Focused History LLM). If the system determines that the patient's chief complaint matches one of approximate100 available symptom-specific surveys (e.g., abdominal pain, cough, dyspnea, headache, or dizziness), the patient receives a survey comprising 10–15 multiple-choice questions. Should the logic connecting to a symptom-specific survey seem inappropriate to the user or if the patient's symptom is not included among the predefined questionnaires, the consultation shifts to the Supplemental History LLM. In this scenario, additional clarifying questions (approximately 10–12) are asked, allowing the AI to collect more nuanced information beyond what is available through the standard templates.

Upon completing these processes, the system integrates all collected data, including relevant medical history, through either the Basic History LLM and Detailed Symptom

Survey or Basic History LLM and Focused History LLM pathway. This integrated approach enables the Diagnosis LLM to leverage both structured survey responses and conversational insights to generate three differential diagnoses. By thoughtfully combining these inputs, the Diagnosis LLM ensures that its diagnostic reasoning considers diverse patient information. Furthermore, by tailoring its workflow to individual patient needs and facilitating multiple opportunities for detailed inquiry, the AI interface synthesizes the available information to guide the formulation of the most clinically appropriate and contextually relevant diagnostic conclusions during early stage evaluations.

*Study Outcomes*

Our primary outcome was the accuracy of the first differential diagnosis. The concordance between the differential diagnoses provided by the physicians and the real-time compound diagnostic medical AI interface with the reference standard was determined by an author (C.H.S., with 2 years of LLM research experience).

Secondary outcomes were as follows: 1) accuracy of the first and second differential diagnoses, 2) time spent on chatting, 3) cost of chatting, and 4) patient satisfaction with care. The costs of chatting and physician labor were calculated based on the average annual salary of residents in South Korea as of 2023. The hourly rate was set at $15.4. Simulated patients were asked to subjectively rate the 13 questions on 4 themes using a 5-point Likert scale (where 1 = strongly disagree, 2 = disagree, 3 = neutral, 4 = agree, and 5 = strongly agree) as follows:

1. Communication with the physician

1) Did the physician interact with you in a kind and empathetic manner? 2) Did the physician

provide clear explanations? 3) Did the physician adequately address your questions? 4) Was the overall conversation comfortable? 5) Were you satisfied with the consultation conducted in a chat format?

2. Physician expertise

1) Did the physician inspire confidence regarding diagnosis and treatment? 2) Were you satisfied with the physician's explanations?

3. Patient-centered care

1) Did the physician respect and consider your opinions? 2) Did the physician adequately explain the treatment options and provide choices?

4. Overall evaluation

1) How satisfied are you with the overall care provided? 2) Would you recommend this physician to others?

*Statistical Analysis*

A chi-square test was performed to compare accuracy between the physicians and real-time compound diagnostic medical AI interface. Repeatability was evaluated based on the proportion of agreement. Statistical analyses were performed using MedCalc version 23.0.2 (MedCalc Software, Mariakerke, Belgium).

**Results**

Three physicians and a real-time compound diagnostic medical AI interface completed 10 clinical vignettes. Each clinical vignette was addressed four times, three times by physicians and once by the chatbot.

*Primary Outcome*

The accuracy of the physicians' first differential diagnosis ranged from 50% (5 out of 10) to 70% (7 out of 10), whereas the real-time compound diagnostic medical AI interface achieved an accuracy of 80% (8 out of 10) (Figure 3A). Although the AI interface had a higher accuracy than that of the physicians, the difference was not statistically significant ($P = 0.17$ and $P = 0.61$). The proportion of agreement for the first differential diagnosis was 0.7 (7 out of 10).

*Secondary Outcome*

The accuracy of the first and second differential diagnoses ranged from 70% (7 out of 10) to 90% (9 out of 10) for physicians, whereas the real-time compound diagnostic medical AI interface achieved an accuracy rate of 100% (10 out of 10) (Figure 3A). Although the AI interface achieved a higher rate than the physicians, the difference was not statistically significant ($P = 0.07$ and $P = 0.32$). Figure 4 presents a representative example of chatting.

Physicians spent an average of between 13 minutes 48 seconds and 21 minutes 30 seconds per clinical vignette, whereas the real-time compound diagnostic medical AI interface spent an average of 9 minutes 17 seconds (Figure 3B). Compared to the physicians' average (16 minutes 46 seconds), the AI interface reduced the average time spent chatting by 44.6%. The average cost of chatting per clinical vignette ranged from 3.3 dollars to 5.4 dollars for physicians, whereas the average cost for the real-time compound diagnostic medical AI interface was 0.08 dollars (Figure 3C). Compared with the physicians' average (4.2 dollars), the AI interface reduced costs by 98.1%.

Patient satisfaction scores from 4.2 to 4.3 for care by physicians ranged, whereas that for the real-time compound diagnostic medical AI interface was 3.9 (Figure 3D). The AI interface's score for patient satisfaction with care was relatively high for two specific items ("Did the physician interact in a kind and empathetic manner?" and "Was the overall conversation comfortable?"), scoring 4.4 and 4.2, respectively. However, the AI interface's score for patient satisfaction with care was relatively low for two specific items ("Did the physician adequately address your questions?" and "Did the physician adequately explain treatment options and provide choices?"), scoring 3.4 and 3.5, respectively.

**Discussion**

We developed an LLM-based real-time compound diagnostic medical AI interface and performed a clinical trial to compare the AI interface with physicians for common internal medicine cases. The accuracy of the first differential diagnosis was within the range of 50–70% for physicians and 80% for the AI interface. Furthermore, the average time for the AI interface (9 minutes 17 seconds) was 44.6% shorter than that of the physicians (16 minutes 46 seconds). The AI interface (0.08 dollars) also reduced costs by 98.1% compared to the physicians' average (4.2 dollars). Patient satisfaction scores ranged from 4.2 to 4.3 for care by physicians and were 3.9 for the AI interface. Therefore, in a clinical trial involving first-time patients in primary care consultations for common internal medicine cases, the AI interface achieved results in less time and at a lower cost than physicians, demonstrating a relatively higher diagnostic accuracy and comparable patient satisfaction.

In a primary care setting based on Q&A interactions with limited information available, the AI interface performed at least on par with general physicians and internal medicine residents. In fact, it demonstrated even higher accuracy in the first differential diagnosis and the first and second differential diagnosis. Furthermore, the proportion of agreement for the first differential diagnosis was relatively high (0.7). In a previous study, the "Articulate Medical Intelligence Explorer" demonstrated significantly higher top-k diagnostic accuracy than primary care physicians across all tested medical specialties, with particularly pronounced improvements in the respiratory and cardiovascular domains ($P < 0.05$).[12]

Compared to the physicians' average time and costs, the chatbot achieved reductions of 44.6% in time and 98.1% in costs, respectively. A previous study demonstrated that chatbots provided substantial time and cost savings, requiring an average of only 5 minutes and $0.21 per case, compared with 50 minutes and $33.24 for radiologists (both $P < .01$).[17] In several countries, primary care appointments are often too short, leaving patients unable to ask all the questions they might have. However, with an AI interface, patients can effectively ask an unlimited number of questions regarding their concerns. Furthermore, given the increasing monetary burden of overall healthcare, the cost effectiveness of AI interfaces can be maximized.

Patient satisfaction scores were slightly lower for the AI interface (3.9) than for the physicians (4.2 to 4.3). In South Korea, patients typically receive brief consultations, which makes it hardly surprising that physicians who devoted an average of 16 minutes and 46 seconds achieved high satisfaction. Given that the AI interface received high scores on questions such as "Did the physician interact in a kind and empathetic manner?" and "Was the overall conversation comfortable?" Compared with physicians under time constraints, the AI interface may demonstrate greater empathy and foster more comfortable conversations when interacting with patients. In a previous study introducing an AI-based chatbot for prostate cancer education,[18] participants who tested the medical chatbot expressed a desire to use one again in the future and supported the integration of chatbots into routine clinical practice. Another previous study revealed that specialists and patient actors consistently rated conversations with the "Articulate Medical Intelligence Explorer" as higher quality than those with primary care physicians, reflecting 'the superior interactions and more effective diagnostic reasoning of the AI.[12] In contrast, in a study examining healthcare workers' perceptions of AI chatbots, participants viewed them as unreliable and believed they should not be used for diagnosis or treatment.[5] However, patients' satisfaction may actually exceed what healthcare providers anticipate. The participants in our study knew that they were interacting with an AI; thus, our findings differ somewhat from those of previous studies.

This study had several limitations. First, the number of clinical vignettes was small, and only a limited number of physicians and simulated patients participated. Second, In LLM research, the risk of stochasticity is well recognized.[19] However, in this study, because the clinical vignettes were formulated as open-ended questions, it was difficult to precisely evaluate stochasticity. Instead, we calculated the proportion of agreement.

In conclusion, in this clinical trial, an LLM-based real-time compound diagnostic medical AI interface demonstrated diagnostic accuracy and patient satisfaction comparable to those of a physician, while requiring less time and lower costs. These findings suggest that LLM-based real-time compound diagnostic medical AI interfaces may have the potential to assist primary care consultations for common internal medicine cases. Such interfaces enable primary care consultations 24 hours a day, 365 days a year, without constraints; therefore, they could be particularly useful for underserved populations or in countries with limited healthcare resources.

**Table 1.** Scores for patient satisfaction with care

| Average score (1 = strongly disagree, 2 = disagree, 3 = neutral, 4 = agree, and 5 = strongly agree) | IM resident (3rd year) | IM resident (2nd year) | General physician | Real-time compound diagnostic medical AI interface |
|---|---|---|---|---|
| **1. Communication with the physician** | | | | |
| 1) Did the physician interact with you in a kind and empathetic manner? | 4.0 | 4.7 | 3.5 | 4.4 |
| 2) Did the physician provide clear explanations? | 4.2 | 4.5 | 4.3 | 4.0 |
| 3) Did the physician adequately address your questions? | 4.6 | 4.5 | 4.1 | 3.4 |
| 4) Was the overall conversation comfortable? | 4.3 | 4.6 | 4.1 | 4.2 |
| 5) Were you satisfied with the consultation conducted in a chat format? | 4.3 | 4.0 | 4.1 | 4.0 |
| **2. Physician's Expertise** | | | | |
| 1) Did the physician inspire confidence regarding diagnosis and treatment? | 4.3 | 4.1 | 4.7 | 3.8 |
| 2) Were you satisfied with the physician's explanations? | 4.6 | 4.1 | 4.2 | 3.7 |
| **3. Patient-Centered Care** | | | | |
| 1) Did the physician respect and consider your opinions? | 4.1 | 4.3 | 3.8 | 4.1 |
| 2) Did the physician adequately explain treatment options and provide choices? | 3.9 | 4.5 | 4.3 | 3.5 |
| **4. Overall Evaluation** | | | | |
| 1) How satisfied are you with the overall care provided? | 4.2 | 4.2 | 4.2 | 3.9 |
| 2) Would you recommend this physician to others? | 4.3 | 4.2 | 4.4 | 4.0 |
| Total average score | 4.3 | 4.3 | 4.2 | 3.9 |

**Figure legends**

**Figure 1.** Overview of clinical trial.

P = physician, CV = clinical vignette, SP = simulated patient

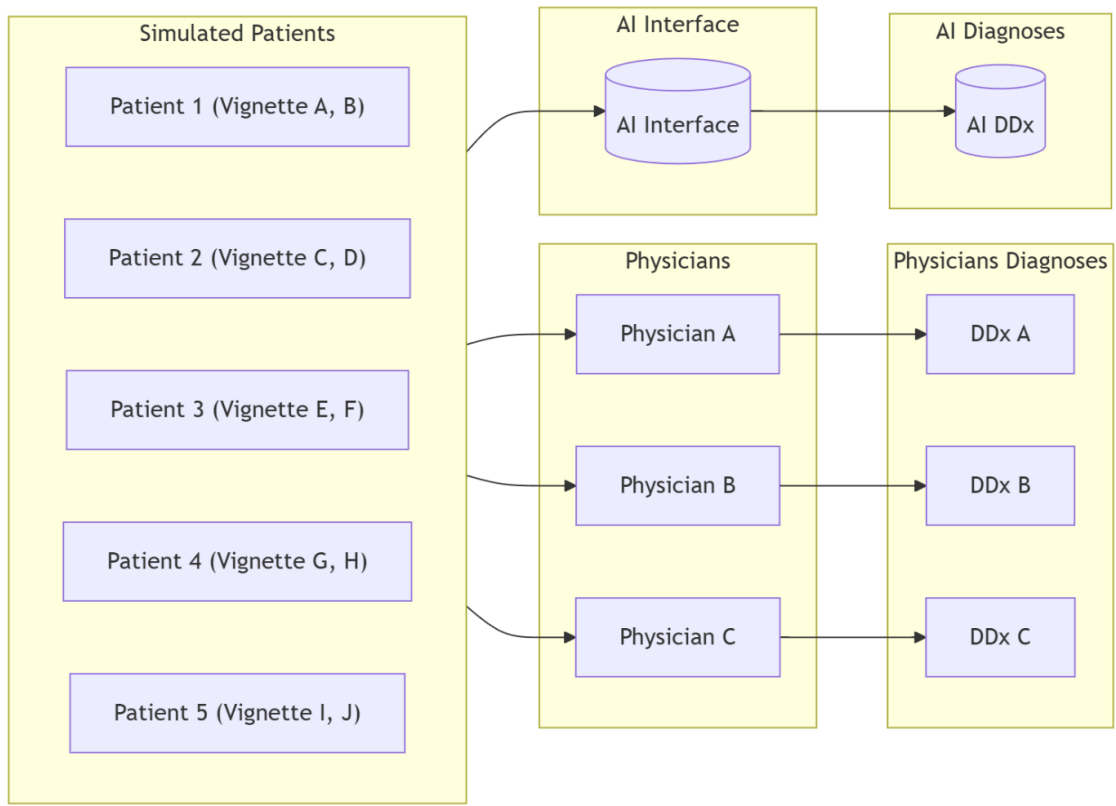

**Figure 2.** Model overview of a large language model-based real-time compound diagnostic medical AI interface

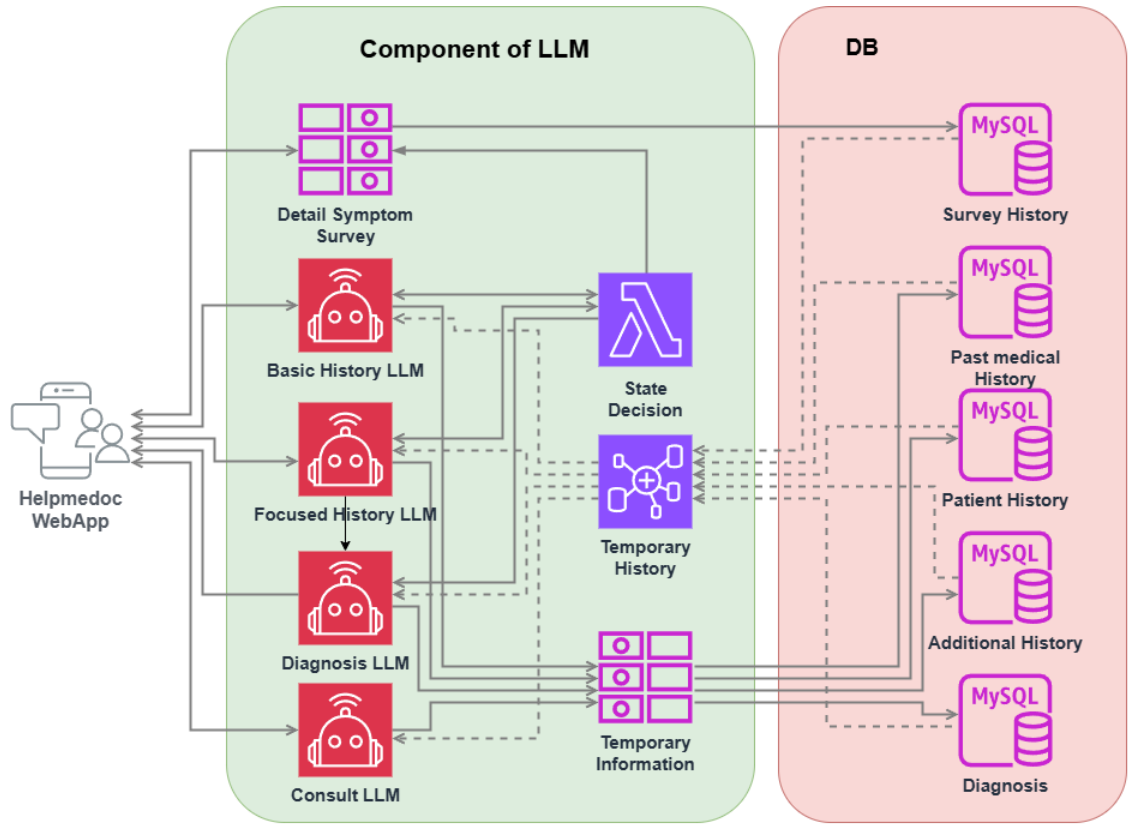

**Figure 3.** Study outcomes. (A) accuracy of the differential diagnosis; (B) time spent chatting; (C) cost of chatting; and (D) patient satisfaction with care

IM = internal medicine

(A)

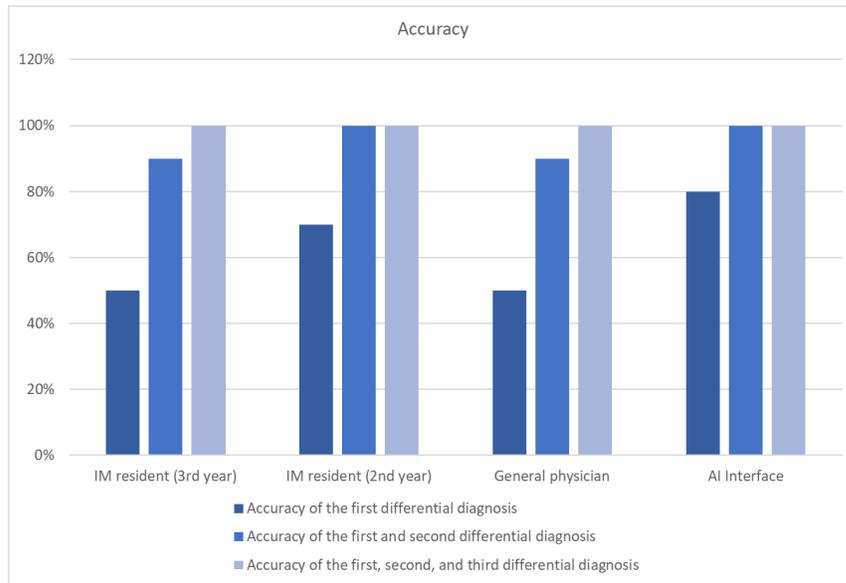

(B)

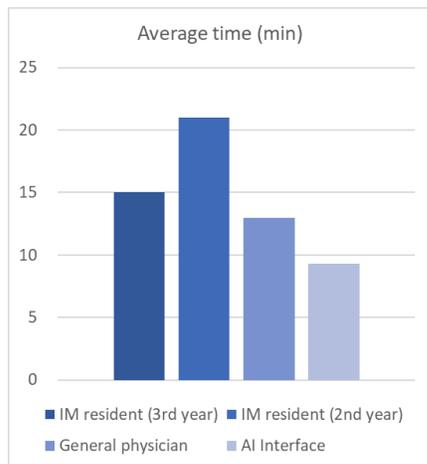

(C)

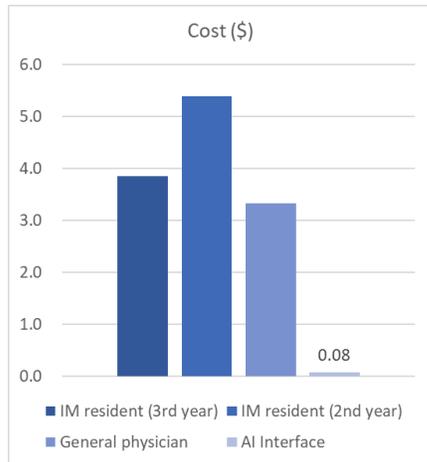

(D)

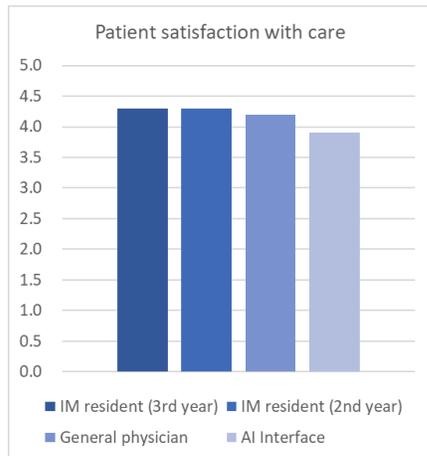

**Figure 4.** Live chat with an LLM-based real-time compound diagnostic medical AI interface for a representative case and its three differential diagnoses.

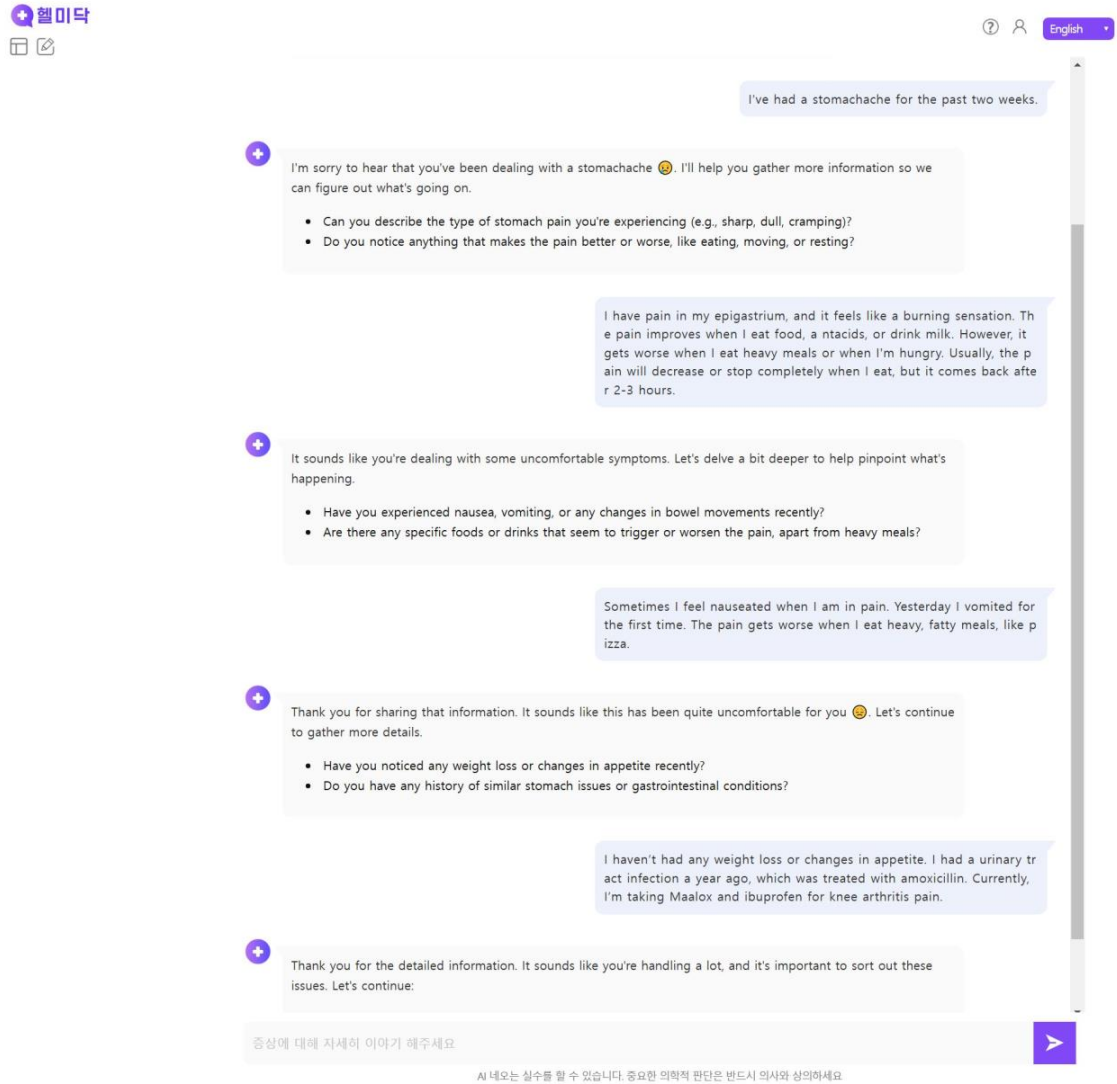

n I eat heavy meals or when I'm hungry. Usually, the pain will decrease or stop completely when I eat, but it comes back after 2-3 hours.

It sounds like you're dealing with some uncomfortable symptoms. Let's delve a bit deeper to help pinpoint what's happening.

- Have you experienced nausea, vomiting, or any changes in bowel movements recently?
- Are there any specific foods or drinks that seem to trigger or worsen the pain, apart from heavy meals?

Sometimes I feel nauseated when I am in pain. Yesterday I vomited for the first time. The pain gets worse when I eat heavy, fatty meals, like pizza.

Thank you for sharing that information. It sounds like this has been quite uncomfortable for you 😊. Let's continue to gather more details.

- Have you noticed any weight loss or changes in appetite recently?
- Do you have any history of similar stomach issues or gastrointestinal conditions?

I haven't had any weight loss or changes in appetite. I had a urinary tract infection a year ago, which was treated with amoxicillin. Currently, I'm taking Maalox and ibuprofen for knee arthritis pain.

Thank you for the detailed information. It sounds like you're handling a lot, and it's important to sort out these issues. Let's continue:

- Have you experienced any heartburn or acid reflux?
- Does the pain wake you up at night?

---

| 진단문 | 설문지 |
|---|---|

- AI Neo is concerned about the following situations and recommends visiting the Gastroenterology department.
- Here is some basic medical information that will be helpful when visiting the hospital:

  1. This could be Peptic Ulcer Disease, especially given the regular use of NSAIDs.
     - Symptoms that match: Epigastric pain, improvement with food/antacids, pain returning 2-3 hours after meals, worsening when hungry, history of NSAID use for arthritis

  2. Chronic Gastritis should be considered due to medication use and symptom pattern.
     - Symptoms that match: Burning sensation in epigastrium, nausea, vomiting, pain pattern related to meals, improvement with antacids, worsening with heavy meals

  3. Functional Dyspepsia might explain the chronic upper abdominal symptoms.
     - Symptoms that match: Upper abdominal pain for extended period, nausea, pain relation to meals, worsening with fatty foods like pizza, no significant weight loss or appetite changes

# Figure 1. Overview of Clinical Trial

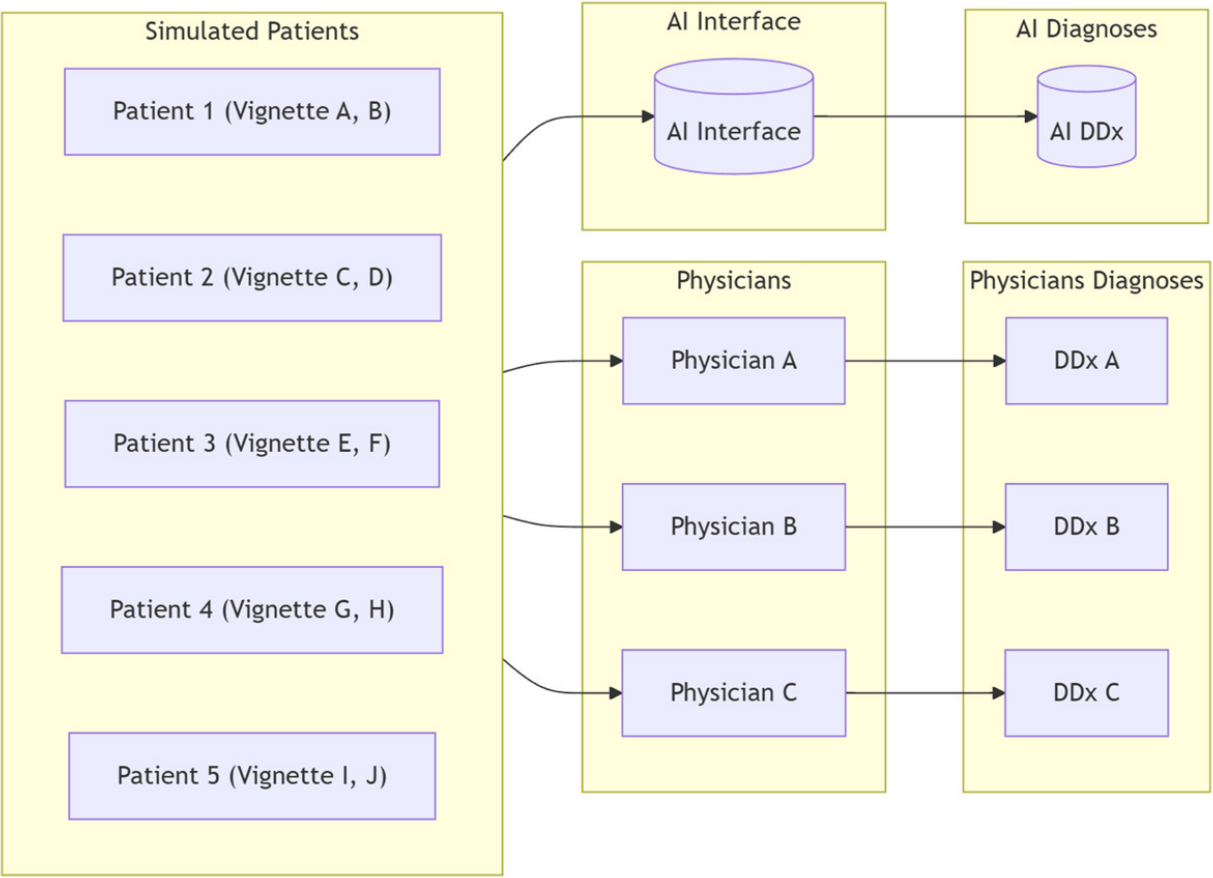

Figure 1: Overview of clinical trial. P = physician, CV = clinical vignette, SP = simulated patient.

# Figure 2. AI Interface Architecture

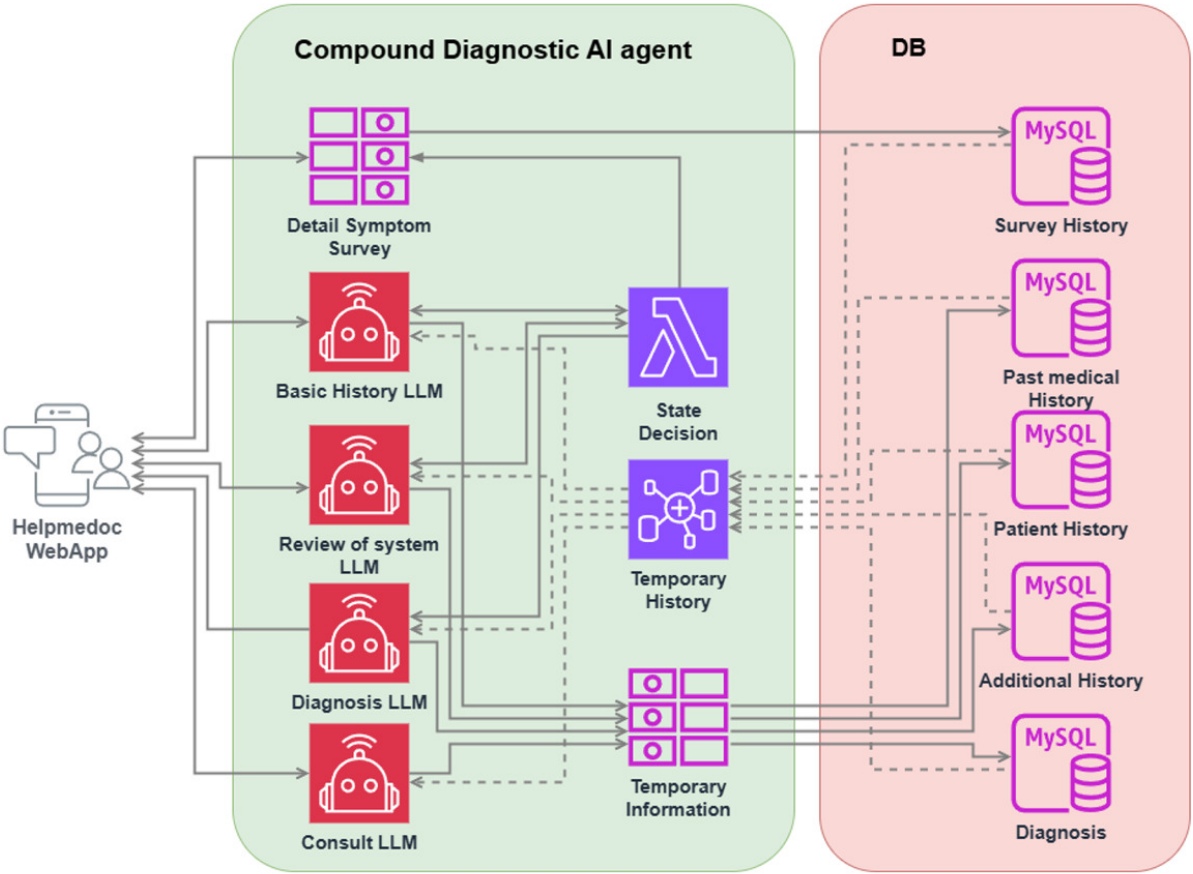

Figure 2: Model overview of a large language model-based real-time compound diagnostic medical AI interface.

# Figure 3. Study Outcomes

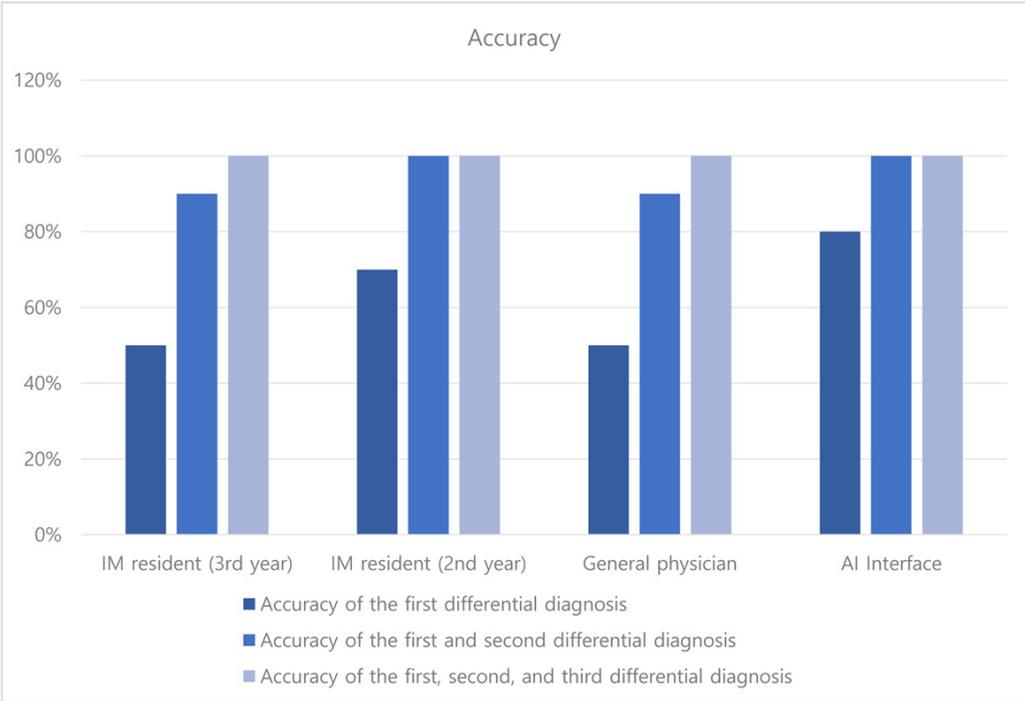

Figure 3: (A) Accuracy of the differential diagnosis. IM = internal medicine.

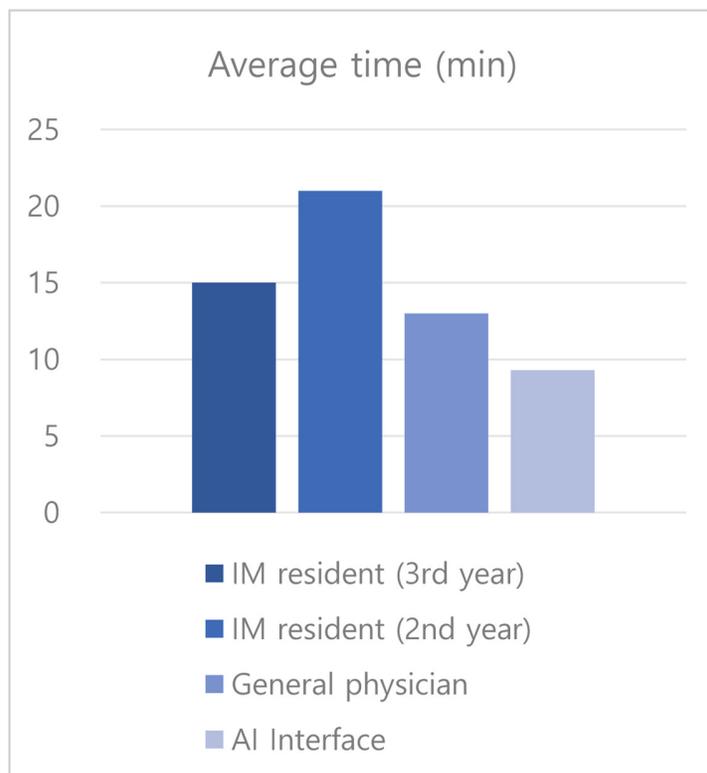

Figure 4: (B) Time spent chatting. IM = internal medicine.

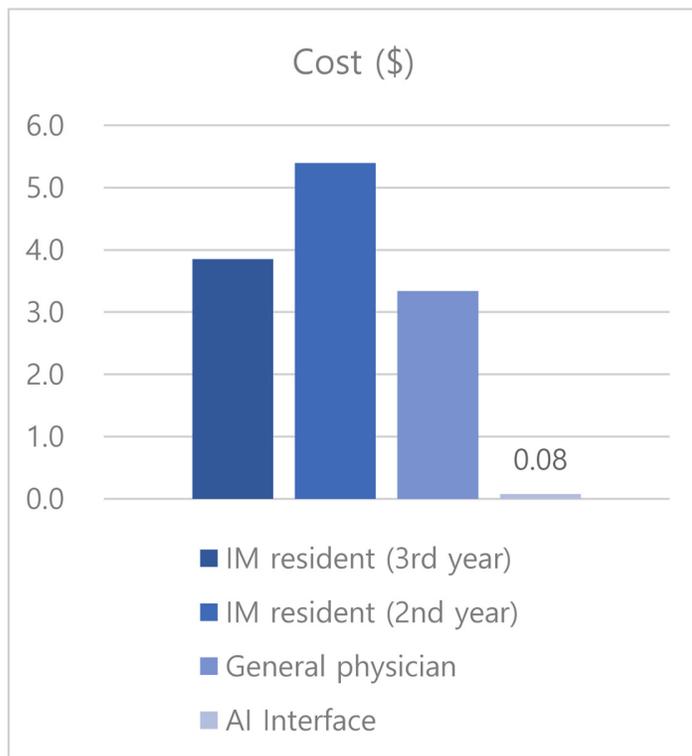

Figure 5: (C) Cost of chatting. IM = internal medicine.

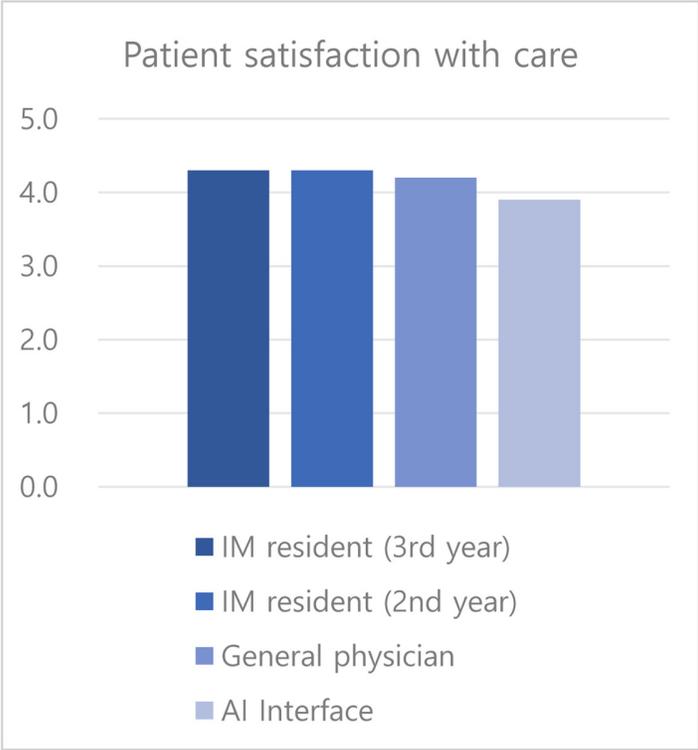

Figure 6: (D) Patient satisfaction with care. IM = internal medicine.

# Figure 4. Live Chat Example

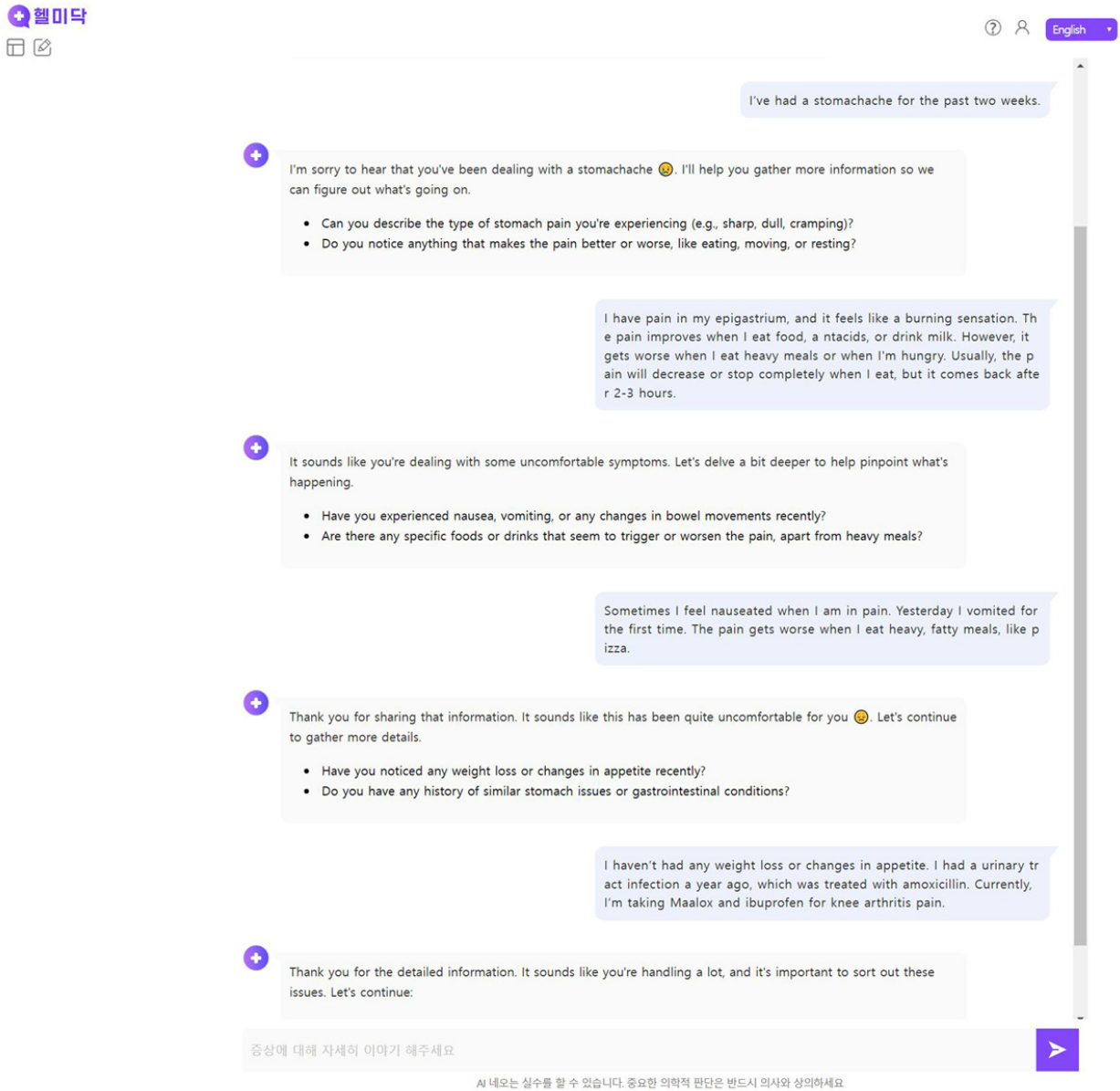

Figure 7: Live chat interface of LLM-based AI interacting with a simulated patient.

Figure 8: AI-generated differential diagnosis summary for the simulated case.